\documentclass{article}

\usepackage[preprint]{neurips_2026}
\usepackage[utf8]{inputenc}
\usepackage[T1]{fontenc}
\usepackage{hyperref}
\usepackage{url}
\usepackage{booktabs}
\usepackage{amsfonts}
\usepackage{amssymb}
\usepackage{nicefrac}
\usepackage{microtype}
\usepackage{xcolor}
\usepackage{graphicx}
\usepackage{amsmath}
\usepackage{enumitem}
\usepackage{booktabs}
\usepackage{tabularx}
\usepackage{array}
\title{A Topological Signature of Learning under Noisy Labels}
\author{Zhongtian Sun}
\author{
Zhongtian Sun$^{1,*}$ \quad
Fan Mo$^{2}$ \quad
Prayag Tiwari$^{3}$ \quad
Kelin Xia$^{4}$ \\[3pt]
$^{1}$University of Kent, United Kingdom \\
$^{2}$University of Oxford, United Kingdom \\
$^{3}$Halmstad University, Sweden \\
$^{4}$Nanyang Technological University, Singapore \\
$^{*}$Corresponding author: \texttt{zs2566@kent.ac.uk}
}
\begin{document}
\maketitle
\begin{abstract}
How does noisy supervision change the representations learned by a neural network? Training accuracy measures agreement with training labels but provides little information about representation geometry. We use persistent homology to track loop structure across distance thresholds in neural activations. Experiments span image classification, text classification, and modular arithmetic. On a fixed CIFAR-10/ResNet-18 training probe, equally sized clean-label and corrupted-label subsets show no detectable persistence difference at epoch~10, whereas corrupted-label subsets have higher persistence at epoch~50. The contrast survives uniform-scale controls, even though predictions match the assigned noisy labels for only 7--11\% of corrupted probe examples at that checkpoint. Normalisation preserves the crossover between clean and noisy training trajectories on CIFAR-10/ResNet-18, whereas CIFAR-100/ResNet-18 shows no crossover under any of the five evaluated constructions. In a separate noise-rate calibration comparison across five pipelines, mean activation norm achieves a higher mean in-sample $R^2$ than raw persistence ($0.936$ versus $0.908$). These findings show that noisy supervision leaves a measurable geometric signature even when agreement with corrupted labels is low. Comparing persistence under scale controls with label-fitting measurements offers a way to study how representations respond to
label corruption beyond what training accuracy reveals.
\end{abstract}

\section{Introduction}
\label{sec:intro}
Neural networks can fit incorrect labels, but training accuracy alone reveals little about how noisy supervision changes their representations. Examples with conflicting labels may occupy nearby regions of representation space, raising the question of how their geometry evolves during training. This question connects to studies of memorization, intrinsic dimension, the information bottleneck, neural collapse, and
grokking \citep{zhang2016understanding,arpit2017closer,ansuini2019intrinsic,
shwartz2017opening,papyan2020prevalence,power2022grokking}. Persistent homology offers a way to examine these changes through loop structure in activation point clouds.

As the pairwise distance threshold increases, points connect, loops form, and those loops eventually fill in. Each loop persists over an interval of thresholds, and total persistence sums the lengths of these intervals. Its interpretation requires attention to activation scale: uniformly enlarging a cloud increases raw persistence while preserving its shape. The analysis therefore combines persistence with activation
scale, normalised point clouds, and geometric measurements that do not use topology. Two questions guide the study: where does the persistence response appear within a noisy training set, and which differences remain after uniform scale is removed?

A fixed CIFAR-10 training probe addresses the first question by comparing equally sized clean-label and corrupted-label subsets within the same trained ResNet-18. No persistence difference is detected at epoch~10, but corrupted subsets have higher persistence at epoch~50 across eight seeds at each of two noise settings. The contrast also appears in summaries invariant to uniform rescaling. At that checkpoint, only
7--11\% of corrupted probe examples are predicted as their assigned noisy label, while 71--79\% are predicted as their original label. The subset comparison thus reveals a geometric difference at a stage when predictions for most corrupted examples still match their original labels.

The broader experiments distinguish estimating noise amount from characterising its effects on representations. Across five pipelines, mean activation norm achieves a higher mean in-sample $R^2$ than raw persistence for noise-rate calibration. Scale-controlled persistence instead measures differences that survive uniform rescaling.
Normalisation preserves clean--noisy crossover on CIFAR-10/ResNet-18, whereas CIFAR-100/ResNet-18 shows no crossover under the five evaluated constructions. An exploratory comparison of random and class-pair relabelling further examines responses to conditions that differ in both realised corruption rate and relabelling structure. These results motivate interpreting persistence in relation to the training stage,
activation scale, and comparison being made.

The contributions are:
\begin{enumerate}
\item \textbf{A training-subset contrast under noisy supervision.}
A fixed-probe comparison identifies higher persistence in corrupted subsets late in training, including under uniform-scale controls, while predictions for most corrupted examples still match their original labels. This provides a group-level view of the response to corruption alongside direct measurements of label fitting.

\item \textbf{An analysis of activation scale and loop structure.}
Normalised point clouds, persistence ratios, and local geometric measurements distinguish differences that survive uniform rescaling from responses sensitive to scale. This clarifies how geometric changes contribute to the observed persistence response.

\item \textbf{Evidence for choosing measurements by task and pipeline.}
Crossover comparisons show that persistence trajectories depend on the pipeline and distance construction. A common calibration protocol finds a stronger average in-sample fit for activation norm than raw persistence, distinguishing the measurement of noise amount from the analysis of its representational effects.
\end{enumerate}

\section{Setup and Observable}
\label{sec:setup}
Let $\mathbf z_i(t)\in\mathbb R^d$ denote the penultimate representation of sample $i$ at epoch $t$, extracted in evaluation mode. The trajectory analysis uses test-set activations to track representation changes on held-out examples during training with noisy labels. These trajectories describe the response on held-out data rather than directly localizing memorization to corrupted training examples. The fixed training probe in Section~\ref{sec:training_probe} compares subsets with clean and corrupted labels within the same trained model. Computing global persistence does not require per-sample corruption labels, whereas the subset comparison uses known corruption membership.

We compute Vietoris--Rips persistent homology with Ripser \citep{bauer2021ripser}, using $\texttt{maxdim}=1$ and no explicit filtration cap. At each distance threshold, the Rips complex connects points within that distance and fills each clique with a simplex. $H_0$ describes connected components, while $H_1$ describes loops that are not filled by triangles. A persistence bar $(b,d)$ records the thresholds at which a feature appears and disappears. Our raw observable is
\begin{equation}
\Phi(t)=\sum_{(b,d)\in\mathcal F_1(t)}(d-b),
\label{eq:phi}
\end{equation}
where $\mathcal F_1(t)$ contains the finite $H_1$ bars whose lifetimes exceed $0.01L_{\max}(t)$, and $L_{\max}(t)$ is the longest finite $H_1$ lifetime at that checkpoint. The sum is zero if no bar is retained. For analyses using this filtered observable, the relative cutoff is fixed across runs, noise levels, and epochs. Some original extensions use unfiltered finite lifetimes; these exceptions are specified in Appendix \ref{app:hyperparams}. The relative cutoff excludes bars with short relative lifetimes without assuming that they are sampling noise. With no explicit filtration cap, feature deaths are not truncated at a manually chosen distance.

Total persistence accounts for both the number of retained loops and their lifetimes, capturing how long each loop persists across distance thresholds rather than counting loops alone. We focus on $H_1$ to study loop structure and compare total persistence with alternative summaries and geometric measurements that do not use topology.

\paragraph{Separating scale and structure.}
The raw analysis uses Euclidean distances between unnormalised activations. Persistence is also computed using unit-normalised activations, cosine dissimilarity, per-feature standardization (z-score), and centered root-mean-square (RMS) rescaling. Unit normalisation and cosine dissimilarity emphasize angular relations, centered RMS rescaling removes a common scale, and z-score standardization adjusts relative feature scales. These are sensitivity analyses, not interchangeable observables. The construction is specified for each result.

Additional summaries include the number of retained $H_1$ bars, $\Phi_{H_1}/\Phi_{H_0}$, and $\Phi/\operatorname{diam}(Z)$, where $\Phi_{H_1}=\Phi$ and $\operatorname{diam}(Z)$ is the maximum pairwise distance under the stated construction. When their denominators are nonzero, both ratios are invariant to uniform positive rescaling, but not to arbitrary anisotropic transformations.

The distinction matters even within a single trained model. Multiplying every activation by the same positive constant multiplies all raw Euclidean birth and death scales by that constant. Since the lifetime cutoff is relative to the longest bar, this operation preserves which bars pass the cutoff while scaling their summed lifetime. An increase in raw $\Phi$ can therefore occur without a change in the ordering of pairwise distances or in the loop structure across corresponding scales. Conversely, changing relative distances can alter persistence after a common scale is removed. The analyses below use normalisation to test sensitivity to representation scale and compare relabelling conditions with their realised corruption rates stated explicitly.

\begin{table}[t]
\centering\small
\caption{Notation. Normalised constructions are named
explicitly when reporting results.}
\label{tab:notation}
\begin{tabular}{@{}l@{\hspace{1.5em}}p{0.70\linewidth}@{}}
\toprule
Symbol & Meaning \\
\midrule
$Z(t)$ & Probe activation cloud at epoch $t$. \\
$\Phi(t)$ & Filtered total finite $H_1$ lifetime on raw
activations, unless qualified. \\
$\Phi_{H_0}$ & Sum of finite $H_0$ lifetimes, excluding
the infinite bar. \\
$\mathcal F_1$, $L_{\max}$ & Retained $H_1$ bars and
maximum finite $H_1$ lifetime, respectively. \\
$\mathrm{sf}$ & Selection fraction for symmetric uniform relabelling \\
$\Delta\Phi$ & Noisy minus clean persistence under
the same construction. \\
$t^\star$ & First sampled checkpoint from which noisy persistence remains above its paired clean reference for all seeds and subsequent checkpoints, after initially lying below it.\\
\bottomrule
\end{tabular}
\end{table}

\paragraph{Sampling and global versus per-class analysis.}
The default global computation uses $n=300$ uniformly sampled activation vectors. Global PH is the primary analysis because it includes cross-class structure. In the classwise protocol, per-class PH is computed only for datasets with at most 20 classes, using classes with at least 10 probe examples and a sampling budget of 50 points per class.
Per-class summaries are summed. This rule was introduced after the original results were available; the original classwise analyses use the separate protocols described in Appendix \ref{app:hyperparams}.

Per-class results use the point counts specified in their appendices, and must not be compared numerically with global sums as if point counts and aggregation were matched. The original experiments used different saved-probe sizes for some extensions; a larger per-class sum alone does not establish a stronger diagnostic.
At $n=300$, the original subsampling study gives rank fidelity $\rho=0.973$ against a 2,000-point reference across 433 configurations, mean subsampling CV $0.022$, and runtime $0.058$\,s per call versus $2.944$\,s at 2,000 points. A separate sensitivity analysis is reported in Appendix~\ref{app:subsampling}.

The original suite spans MNIST \citep{lecun1998gradient}, FashionMNIST \citep{xiao2017fashion}, CIFAR-10/100 \citep{krizhevsky2009learning}, AG News \citep{zhang2015character}, Tiny-ImageNet \citep{deng2009imagenet}, ImageNette \citep{Howard_Imagenette_2019}, and modular arithmetic \citep{power2022grokking}. Architectures include MLPs, ResNets \citep{he2016deep}, ViTs \citep{dosovitskiy2020image}, DistilBERT \citep{sanh2019distilbert}, and small Transformers \citep{vaswani2017attention}. Noise conditions include symmetric uniform relabelling, Patrini class-pair relabelling, cyclic asymmetric noise, a pixel-distance-based instance-dependent construction in the ViT experiment, and CIFAR-10N human annotations \citep{patrini2017making, wei2021learning}. The intended instance-dependent condition for ResNet-18 is analysed as a random-relabeling condition in Section \ref{sec:crossover}.

The original suite reports over 900 training runs and 8,000 PH computations. The normalisation reanalysis covers 1,540 saved activation files. These counts refer to different stages of the study, not additional independent training runs. Hyperparameters and experiment-specific seed counts appear in the appendices.

\section{Early Response and Late Training-Set Evidence}
\label{sec:complementarity}
At epoch 10, raw persistence is associated with noise level in the evaluated MLP, ResNet-18, ViT-Tiny, and DistilBERT pipelines (Appendix~\ref{app:fading}). In the separate five-pipeline calibration comparison, mean activation norm achieves a higher mean in-sample $R^2$ than raw persistence, 0.936 versus 0.908 in Table \ref{tab:noise_compare}. Section~\ref{sec:crossover} examines sensitivity to activation scale through normalisation. 


The association varies over training and across pipelines. On CIFAR-10/ResNet-18, the correlation between noise level and raw persistence changes from $\rho=-0.969$ at epoch 10 to $-0.225$ at epoch 50. On MNIST/MLP, it changes from $-0.931$ to $-0.318$, while ViT-Tiny/CIFAR-10 retains a correlation of $-0.962$ at both epochs (Appendix~\ref{app:fading}). These correlations describe the ordering of noise conditions across trained models. A weaker correlation does not by
itself establish a crossover between clean and noisy conditions or locate memorization within the training set. The following experiment compares clean and corrupted subsets within the same trained model.

\subsection{A fixed training probe}
\label{sec:training_probe}


The fixed probe contains 4,000 CIFAR-10 training images, with 400 sampled from each true class using RandomState(12345). The same images are evaluated without augmentation across seeds, conditions, and checkpoints. Under symmetric uniform relabelling at $sf=0.25$ and $0.5$, corruption membership identifies labels that actually changed. The clean and corrupted groups are matched in size, but not in class composition, and PH is computed on 300 points from each group.
Each condition uses eight seeds. Paired comparisons use the exact two-sided Wilcoxon signed-rank test.

At epoch~50, corrupted subsets have higher raw persistence and higher values of both scale-invariant ratios (Table~\ref{tab:training_probe}). The corrupted/clean ratios for raw persistence are $1.29$ and $1.37$ at the two noise settings. Each summary is higher for corrupted subsets in all eight seeds, except $\Phi/\operatorname{diam}(Z)$ at $\mathrm{sf}=0.25$, where it is higher in seven seeds.

\begin{table}[t]
\centering\small
\caption{Corrupted/clean persistence ratios at epoch 50 on the fixed CIFAR-10/ResNet-18 training probe, with eight seeds per noise setting. Each entry is the median corrupted-subset statistic divided by the median clean-subset statistic across seeds. Corrupted subsets have larger values in all eight seeds except for diameter-normalised persistence at $sf=0.25$, where the direction agrees in seven seeds. Exact two-sided paired Wilcoxon tests give $p=0.0078$ for the
eight-of-eight comparisons and $p=0.0156$ for the seven-of-eight comparison.}
\label{tab:training_probe}
\begin{tabular}{@{}ccccc@{}}
\toprule
$\mathrm{sf}$ & Raw $\Phi$ & $\Phi_{H_1}/\Phi_{H_0}$ & $\Phi/\operatorname{diam}(Z)$ & Loop count \\
\midrule
0.25 & 1.29 & 1.23 & 1.44 & 1.24 \\
0.50 & 1.37 & 1.28 & 1.46 & 1.36 \\
\bottomrule
\end{tabular}
\end{table}

\begin{figure}[h]
\centering
\includegraphics[width=\linewidth]{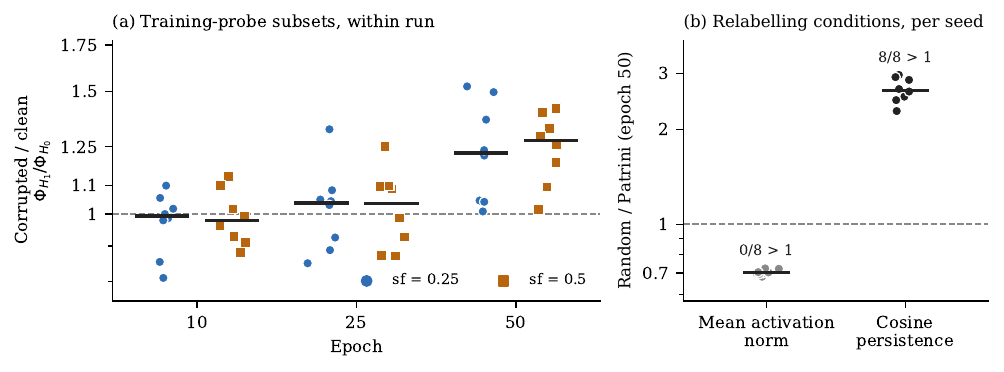}
\caption{Seed-level persistence comparisons.
(a) Corrupted/clean ratios of $\Phi_{H_1}/\Phi_{H_0}$ on the fixed CIFAR-10/ResNet-18 training probe, with eight seeds per setting. Dots show within-run ratios and horizontal bars show their medians. These differ from the ratios of seed medians reported in Table~\ref{tab:training_probe}.
(b) Epoch-50 random/Patrini ratios for mean activation norm and cosine persistence. Each dot shows one seed's median ratio across three nominal settings. The random arm changes approximately 1.8 times as many labels as the Patrini arm, so the comparison includes differences in both corruption rate and relabelling structure. Both panels use logarithmic vertical axes. Dashed lines mark a ratio of one.}
\label{fig:main_evidence}
\end{figure}

No difference between clean and corrupted subsets is detected at epoch~10. At epoch~25, only loop count differs, and only at $\mathrm{sf}=0.25$. By epoch~50, a persistence contrast is present, although only 7--11\% of corrupted probe examples are predicted as their assigned noisy label and 71--79\% as their original label. The persistence contrast is therefore present at a checkpoint where predictions for most corrupted examples still match their original labels. Across six epoch-wise comparisons, test-set and training-probe statistics show no statistically detectable difference (minimum $p=0.11$). These comparisons concern global probe statistics and do not establish equivalence. The excess persistence relative to separately fitted class-conditional Gaussian nulls does not differ detectably between the clean and corrupted subsets (all $p\geq0.25$). Because each null is fitted using its subset's covariance, this control incorporates second-order class geometry; its negative result limits claims that the effect uniquely identifies a deformation beyond that geometry. Figure~\ref{fig:main_evidence}(a) shows the seed-level subset ratios across checkpoints.

\section{Geometric Interpretation and Controlled Comparisons}
\label{sec:folding}

Conflicting labels can reshape local neighborhoods in representation space as a network learns to separate nearby examples. Folding is one possible geometric interpretation of this process. Persistent homology measures loop structure in the sampled activation cloud, but persistence also depends on sampling, class geometry, and distance scale. The following controlled comparisons examine how the placement of label contradictions affects persistence, while the training-probe and normalisation analyses help distinguish these effects.

\paragraph{Scope of the motivating model.}
Networks composed of affine maps, convolutions, ReLU activations, average or max pooling, and residual addition are piecewise affine. BatchNorm in evaluation mode is a fixed affine map and preserves this property. This argument applies to the corresponding operations in the residual networks studied here, but not to LayerNorm or softmax attention. Piecewise-affine structure alone guarantees neither an $H_1$ cycle nor a lower bound on its lifetime.

\paragraph{Class-pair controls.}
The next comparison varies which classes receive conflicting labels, with class pairs selected solely by centroid distances in clean representations. The NEAR and FAR conditions each contain five disjoint pairs, covering all ten CIFAR-10 classes once per condition. Their mean centroid distances are $3.32$ and $4.94$, respectively. With ten seeds per condition and a direction of comparison specified in advance, raw persistence at epoch~50 is $1.445$ times as large in FAR as in NEAR. The direction agrees in every seed ($p=0.002$). Thus, the placement of corruption across classes affects the raw persistence response, which remains sensitive to activation scale.

An analysis across class pairs relates centroid distance to the noisy/clean persistence ratio. At $sf=0.5$, clean centroid distances and noisy/clean ratios are averaged across seeds for each pair before computing the correlation
across pairs. At epoch~50, the Spearman correlations are $0.867$ for CIFAR-10/ResNet-18, $0.614$ for CIFAR-10/ViT-Tiny, and $0.413$ for the 60 evaluated CIFAR-100/ResNet-18 pairs. These associations vary over training, as detailed in Appendix~\ref{app:pair_controls}. The appendix also retains the original exploratory experiment, in which pair selection partly used persistence.



\begin{figure}[t]
\centering
\includegraphics[width=\linewidth]{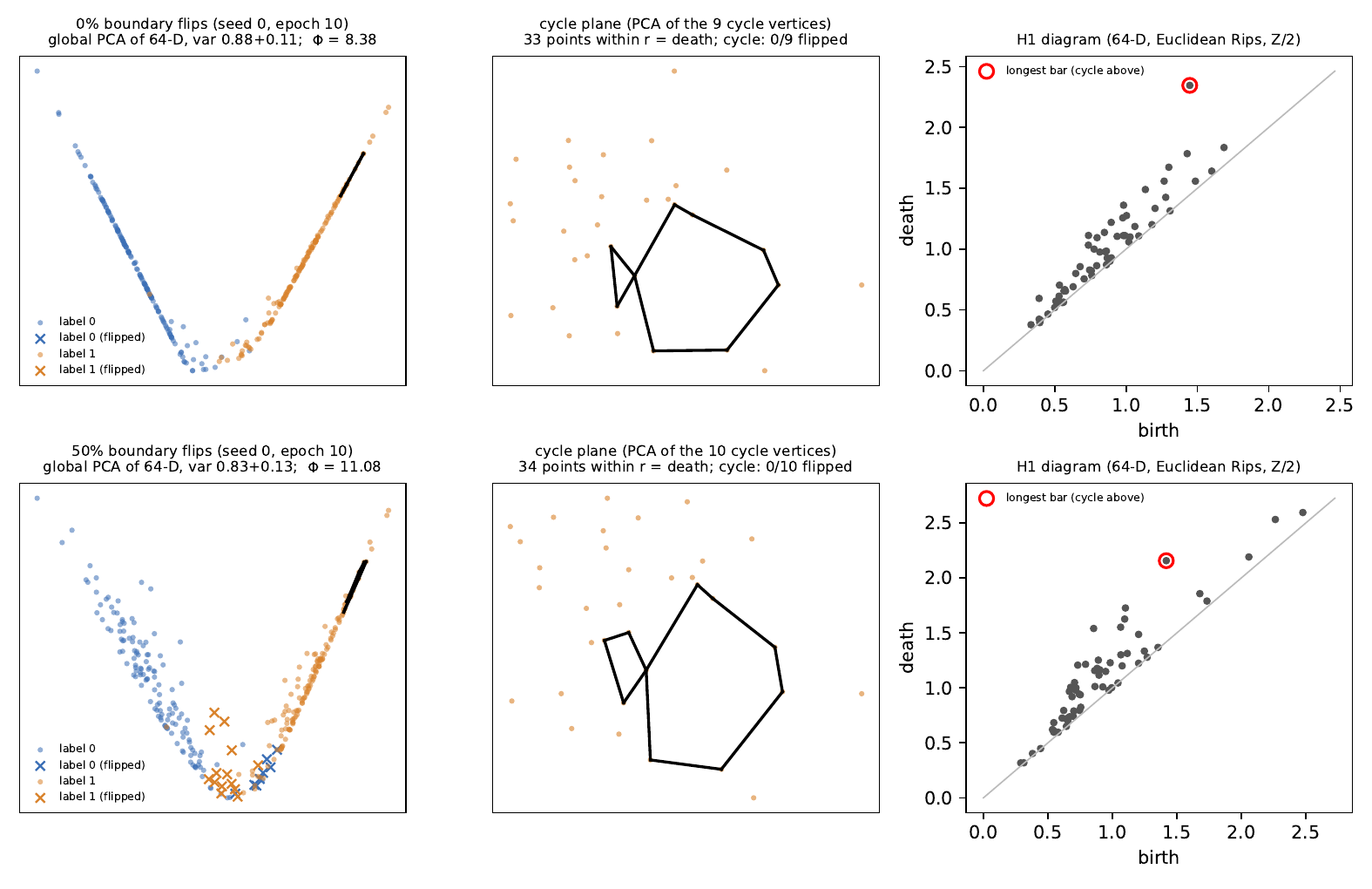}
\caption{Representation and persistent cycles in the synthetic two-Gaussian task, using seed 0 at epoch 10.
The top and bottom rows show 0\% and 50\% of near-boundary labels flipped, respectively; the boundary region is $|x_0|<1$. Left: PCA projections of the 300-point subsamples of 64-dimensional activations, fitted separately for each condition. Colour denotes the observed label and crosses mark flipped labels. Black edges show a representative of the longest finite $H_1$ bar, computed in the original 64-dimensional Euclidean Rips filtration over $\mathbb{Z}/2\mathbb{Z}$. Middle: the same representative projected onto its own first two principal components, with subsample points lying within its death radius of a cycle vertex. Right: persistence diagrams with the represented bar circled. Total persistence is 8.38 and 11.08, respectively. Neither displayed representative contains flipped points.
Across 20 seeds, the longest-bar representatives show no detected enrichment for flipped points.}
\label{fig:synthetic_cycle}
\end{figure}

\paragraph{Synthetic contradictions and local geometry.}
The two-dimensional experiment uses two Gaussian clusters with label flips near the boundary (Appendix~\ref{app:hyperparams}). Across 140 individual seed runs with 0--50\% of near-boundary labels flipped, the association between flip rate and persistence at epoch~10 is $\rho=0.33$ ($p=6.1\times10^{-5}$). Averaging within each of the seven flip rates gives $\rho=1.000$ (exact two-sided $p=0.0004$). The first analysis captures variation across individual runs, while the second describes the ordering of the rate means. Figure~\ref{fig:synthetic_cycle} shows the activation projections and representative persistent cycles for seed 0 at epoch 10. Total persistence is higher under boundary-label corruption, but the longest-bar representative contains no flipped points in either condition.

Local geometric measurements provide a complementary view. On CIFAR-10/ResNet-18, local PCA dimension with $k=10$ correlates with noise level at $\rho=0.946$ at epochs 25 and 50 when the fully randomised condition is excluded ($sf\in\{0,0.25,0.5\}$, five seeds). In the exploratory ResNet-18 relabelling comparison, local PCA dimension is higher in the random-relabeling arm in all 24 seed--rate pairs at epoch 50. These pairs comprise eight seeds at three settings,
and the arms differ in realised corruption rate as well as relabelling structure. Multiplying activations by $9.3$ changes the geodesic-to-Euclidean measure by less than $10^{-9}$. Local dimension measures how many directions contribute to neighborhood variation, while persistence describes loop structure across distance thresholds. These measurements characterize the sampled representations without uniquely identifying folding or establishing a new topological hole in the underlying data manifold.

\section{Normalisation, Crossover, and Noise Type}
\label{sec:crossover}

\paragraph{Crossover across distance constructions.}

We report the first persistent crossing: the earliest sampled checkpoint from which noisy persistence remains above its seed-matched clean reference at every subsequent checkpoint for all evaluated seeds, after initially lying below it. On CIFAR-10/ResNet-18 at $sf=0.5$, this occurs at epoch 20 under cosine, unit normalisation, centered RMS, and z-score constructions, compared with epoch 30 under raw Euclidean distance. On MNIST/MLP, all five seeds cross at epoch 30 under raw, unit-normalised, and cosine constructions, while no seed crosses under RMS or z-score (Table~\ref{tab:crossover_reanalysis}). FashionMNIST shows no crossover under any of the five constructions. These results depend on both the pipeline and the distance construction.

Two trajectory properties help describe these cases: clean persistence declines (C1), and noisy persistence declines less rapidly (C2). For crossover to occur, the initially negative gap must close within the observed horizon. C1 and C2 describe the measured trajectories rather than provide a general crossover criterion, and C2 requires observations from noisy training. In the held-out ResNet-34/CIFAR-10 experiment, the clean trajectory supported a prospective prediction of no crossover, subsequently confirmed in six noisy runs against three clean reference runs (Appendix~\ref{app:r34}). The recomputed CIFAR-100/ResNet-18 analysis shows no crossover in any of the three seeds under any of the five constructions. Under unit normalisation, the epoch-50/epoch-10 persistence ratios are $0.843$ for clean training and $0.434$ for noisy training. Persistence declines in both cases, with a larger proportional decline under noisy supervision. Crossover timing refers to sampled checkpoints and does not identify when individual corrupted examples begin to be memorized.

\begin{table}[tbp]
\centering\small
\caption{First persistent crossing under the evaluated constructions: the earliest sampled checkpoint from which all seed-paired noisy trajectories remain above their clean references. ``None'' denotes no persistent crossing over the evaluated checkpoints. The CIFAR-10/ResNet-18 comparison uses $sf=0.5$.}
\label{tab:crossover_reanalysis}
\begin{tabular}{@{}lclll@{}}
\toprule
Pipeline & Seeds & Raw & Unit / cosine & RMS / z-score \\
\midrule
CIFAR-10 / ResNet-18 & 5 & 30 & 20 & 20 \\
MNIST / MLP & 5 & 30 & 30 & None \\
CIFAR-100 / ResNet-18 & 3 & None & None & None \\
\bottomrule
\end{tabular}
\end{table}

\begin{figure}[tbp]
\centering
\includegraphics[width=0.94\textwidth]{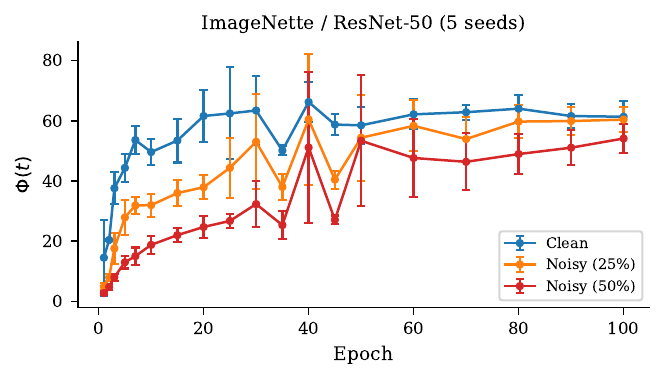}

\caption{Raw persistence on ImageNette/ResNet-50 over 100 epochs. Curves show the mean across seeds, with error bars indicating the sample standard deviation (SD, ddof = 1).}
\label{fig:imagenette}
\end{figure}

\paragraph{Persistent clean trajectories.}
On ImageNette/ResNet-50, the mean clean trajectory rises overall over 100 epochs and remains above the noisy means at the displayed checkpoints (Figure~\ref{fig:imagenette}). The curves fluctuate during training, showing that persistence can change without crossover. The ResNet-34, ViT-Tiny, and Tiny-ImageNet experiments provide further comparisons, although differences in datasets, sampling, and normalisation preclude a general rule based on capacity or depth.

\paragraph{A direct scale check.}
Under uniform positive rescaling of a fixed geometry, $\Phi_{H_1}$ and $\Phi_{H_0}$ scale by the same factor, giving a slope of one when $\log\Phi_{H_1}$ is plotted against $\log\Phi_{H_0}$. On CIFAR-10/ResNet-18 at epoch~50, the estimated slope is $1.577$ (95\% interval $[1.057,2.052]$) using four noise levels and $2.260$ ($[1.903,2.646]$) after excluding the fully randomized level. Across four vision configurations and 14 checkpoints excluding CIFAR-100, slopes differ from one in 52/56 cells after Benjamini--Hochberg false-discovery-rate correction using four levels, compared with 19/56 using three. The latter analysis has fewer points per slope estimate. The reported ViT-Tiny slope is much closer to one at $1.029$ ($[1.008,1.055]$). These results reject pure uniform rescaling in the stated comparisons, while leaving other geometric explanations open.



\paragraph{Exploratory comparison of relabelling conditions.}
The ResNet-18 experiment compares a Patrini asymmetric condition with a condition originally intended to be instance-dependent. In the latter, entropy scores were computed using a shuffled data loader and then assigned to dataset indices without restoring their order. The resulting selection is therefore treated as effectively random, followed by uniform relabelling. We refer to it as the random-relabeling arm. At nominal settings of 0.1, 0.2, and 0.4, the realised changed-label fractions are 0.050, 0.100, and 0.200 for the Patrini arm, and 0.089--0.090, 0.179--0.182, and 0.359--0.361 for the random arm. Thus the conditions differ in both corruption amount and relabelling structure.

\begin{table}[tbp]
\centering\small
\caption{Random versus Patrini asymmetric relabelling on ResNet-18. The conditions have unequal realised corruption rates, as stated in the text. Each seed contributes the median of its three per-setting random/Patrini ratios. Entries give the number of seeds agreeing with the indicated direction out of eight, with exact two-sided Wilcoxon $p$-values in parentheses. Arrows indicate the random arm relative to the Patrini arm, not changes over training.}
\label{tab:noise_seed_controls}
\begin{tabular}{@{}lccc@{}}
\toprule
Observable / direction & Epoch 10 & Epoch 25 & Epoch 50 \\
\midrule
Raw $\Phi$ & $\downarrow$ 8/8 (.0078) & $\downarrow$ 8/8 (.0078) & $\uparrow$ 8/8 (.0078) \\
Cosine $\Phi$ ($\uparrow$) & 7/8 (.0156) & 8/8 (.0078) & 8/8 (.0078) \\
$\Phi_{H_1}/\Phi_{H_0}$ ($\uparrow$) & 7/8 (.0156) & 6/8 (.0391) & 8/8 (.0078) \\
Mean norm ($\downarrow$) & 8/8 (.0078) & 8/8 (.0078) & 8/8 (.0078) \\
\bottomrule
\end{tabular}
\end{table}


For each observable and seed, we take the median of the three per-setting random/Patrini ratios. Table 4 reports exact two-sided Wilcoxon tests of these seed-level ratios against one, using eight seeds. Raw persistence is lower in the random arm at epochs 10 and 25 and higher at epoch 50 in all eight seeds. Cosine persistence is higher in seven, eight, and eight seeds, respectively, while mean activation norm is lower in all eight seeds at each checkpoint (Table~\ref{tab:noise_seed_controls}). The $\Phi_{H_1}/\Phi_{H_0}$ ratio is higher in seven, six, and eight
seeds. At epoch 50, the ratio of pooled median cosine persistence is 2.759. Figure~\ref{fig:main_evidence}(b) shows the seed-level epoch-50 comparison. A pooled analysis of 48 runs gives a partial Spearman correlation of 0.472 between the relabelling-arm indicator and cosine persistence after controlling for mean norm and diameter ($p=0.0007$). These observations comprise two arms and three settings for each of eight seeds; the pooled significance calculation does not account for within-seed dependence and is descriptive. Neither this analysis nor the paired comparison isolates noise structure from corruption amount. On CIFAR-10N, unit-normalised and cosine persistence correlate with corruption rate at $\rho=0.936$ and $0.929$, respectively, compared with $\rho=0.302$ ($p=0.15$) for raw persistence. Across these experiments, activation scale and normalised persistence describe different aspects of the response to label noise.

\paragraph{Alternative summaries and sampling.}
Loop count is invariant to uniform positive rescaling under the relative lifetime cutoff. Its reported noise correlations are $\rho=-0.799$ on CIFAR-10/ResNet-18 and $-0.830$ on CIFAR-10/ViT-Tiny (both $p<0.001$), but are not significant on the reported CIFAR-100, FashionMNIST, and MNIST comparisons (Appendix~\ref{app:pair_controls}). Thus the number of retained loops captures part of the response, while their lifetimes retain additional descriptive detail. Separately, the added CIFAR-10/ResNet-18 epoch-10 sampling analysis gives $\rho=-0.850$ at 100 points and $-0.945$ at each tested size from 200 to 2,000 points. Equal rank correlations do not imply equal persistence values or eliminate finite-sampling effects.

\section{Grokking and Decreasing Persistence}
\label{sec:grokking}


Grokking refers to delayed generalization after training accuracy has saturated \citep{power2022grokking}. We examine twelve runs of two-layer Transformers trained on modular addition and multiplication with primes 97 and 113, using three seeds per task-prime combination. Raw persistence decreases across the grokking transition in ten runs but increases by $3.7$ and $4.5$ in the other two (Figure~\ref{fig:grokking}; Appendix~\ref{app:grokking}). The mean change is $-35.2$ across all twelve runs and $-43.1$ among the ten decreasing runs. We define $\Delta\Phi_{\mathrm{grok}}$ as the difference in mean test-set persistence between checkpoints with training accuracy $\geq 0.99$ and test accuracy $\geq 0.90$, and those with training accuracy $\geq 0.99$ and test accuracy $< 0.50$. Persistence is computed on 300 test points as the sum of all finite $H_1$ lifetimes, without a relative cutoff. The results associate delayed generalization with lower raw persistence in most runs, extending the analysis beyond label corruption. However, they do not establish a causal role for unfolding. The evidence remains limited to modular arithmetic, and the observed pattern has not been shown to hold under scale normalisation.



\section{Diagnostic Use: Calibration and Representation Comparisons}
\label{sec:tools}


An early representation statistic can be calibrated against the nominal noise setting within a fixed pipeline.
Calibration requires runs at known settings, including a clean reference. A clean run alone cannot determine the slope, although applying the calibration to a new probe does not require individual corruption labels. At epoch 10, we fit a separate ordinary least-squares regression with an intercept within each of five pipelines, predicting the nominal noise setting from each representation statistic over individual runs. Table~\ref{tab:noise_compare} reports in-sample $R^2$, not held-out prediction performance. Mean activation norm achieves a mean $R^2$ of 0.936, compared with 0.908 for raw persistence. Excluding the fully randomised condition gives 0.892 and 0.729, respectively. Appendix~\ref{app:noise_est} specifies the pipelines and calibration protocol.


\begin{table}[t]
\centering\small
\caption{In-sample noise-rate calibration across five pipelines at epoch 10. Within each pipeline, ordinary least squares with an intercept predicts the nominal noise setting from each statistic over individual runs. Ranges and means summarize $R^2$ across pipelines; ranges are not confidence intervals.}
\label{tab:noise_compare}
\begin{tabular}{@{}lcc@{}}
\toprule
Observable & $R^2$ range & Mean $R^2$ \\
\midrule
Mean activation norm & 0.890--0.990 & 0.936 \\
Raw $\Phi$ & 0.871--0.927 & 0.908 \\
Point-cloud diameter & 0.566--0.972 & 0.790 \\
Unit-normalised $\Phi$ & 0.382--0.821 & 0.606 \\
$\Phi_{H_1}/\Phi_{H_0}$ & 0.115--0.757 & 0.500 \\
\bottomrule
\end{tabular}
\end{table}


Normalised persistence describes representation differences after specified aspects of activation scale are removed. The within-model subset comparison and the normalised trajectories address this descriptive role. Population-level calibration is a separate task from detecting individual corrupted examples or improving robust classification accuracy; its regression $R^2$ should not be interpreted as detection or classification performance. Two exploratory experiments examine whether persistence can guide training decisions. The checkpoint-selection procedure uses validation accuracy to choose between the persistence-peak checkpoint and the final checkpoint (Appendix~\ref{app:tgcs_algo}). It therefore uses validation labels and does not establish an advantage over selecting the best checkpoint by validation accuracy. Adaptive label smoothing improves on fixed smoothing at one tested noise level, but its displayed mean accuracy is $0.07$ percentage points lower.

Additional results show why persistence should not be treated as an objective to maximize or minimize (Appendix~\ref{app:negative}). A covariance effective-rank proxy correlates with raw persistence on CIFAR-10/ResNet-18 ($\rho=0.831$), but shows little association on FashionMNIST ($\rho=0.002$). Persistence-peak height has only a modest overall correlation with final accuracy ($\rho=0.27$). Label smoothing and weight decay reduce persistence, whereas topology-guided training increases it, without a corresponding monotone ordering of test accuracy. Persistence thus characterizes representation changes, while its value for choosing training interventions requires separate evaluation.

\section{Scope and Limitations}
\label{sec:scaling}

The results vary across datasets and architectures. AG News/DistilBERT shows a positive early noise association, whereas the evaluated vision settings show negative associations. Clean persistence remains elevated on ImageNette and Tiny-ImageNet, while the recomputed CIFAR-100/ResNet-18 trajectories show no crossover. These differences make calibration specific to each pipeline. Understanding the role of depth requires comparisons with matched data and measurement protocols. Sampling also matters. From epochs~10 to~50, clean global persistence on ViT-Tiny/CIFAR-10 rises from $32.1$ to $66.1$ at $n=300$, and the classwise sum rises from $260.3$ to $403.4$. Clean global persistence on Tiny-ImageNet/ResNet-18 rises from $84.4$ to $165.6$ (Appendices~\ref{app:vit} and~\ref{app:tinyimagenet}). Classwise PH measures structure within classes, while global PH also includes relations between classes. On Tiny-ImageNet, these analyses use 300 points per class across 50 classes and 300 points overall, respectively. Their different budgets and aggregation rules must be accounted for when comparing discrimination.

The late training-probe contrast distinguishes clean and corrupted groups at a checkpoint where most corrupted examples
are predicted as their original labels.  The class-conditional Gaussian control detects no group difference in excess persistence, leaving the interpretation of folding to be assessed alongside class geometry, anisotropy, and sampling.
In the exploratory ResNet-18 relabelling comparison, noise structure and realised corruption rate vary between arms, leaving their individual contributions to the persistence contrast unresolved. Several questions remain open. Sensitivity to the fixed cutoff of $0.01L_{\max}$ requires a threshold sweep, and finer checkpoint sampling would improve the timing resolution of the 14-checkpoint study. The current summaries describe groups of examples, with classwise analysis additionally requiring class labels. Extending the piecewise-affine account to LayerNorm and attention, and testing normalised persistence during grokking beyond modular arithmetic, remain directions for further work.

\section{Related Work}
\label{sec:related}

Persistent homology has been used to study neural representations and network structure \citep{naitzat2020topology,rieck2018neural,hofer2017deep}, filter noisy labels and regularize classifiers \citep{wu2020topological,chen2019topological}, and investigate intrinsic dimension and generalization \citep{birdal2021intrinsic}. This work examines how persistence evolves during training with noisy labels, using comparisons between clean and corrupted training subsets and controls for activation scale. The ability of neural networks to fit random labels, and their tendency to learn patterns before memorizing noise, are well established \citep{zhang2016understanding,arpit2017closer}. Risk correction and sample filtering seek to improve learning under label noise \citep{natarajan2013learning,li2020dividemix}. Our study uses persistence to characterize the accompanying representation changes at the population level.

Other accounts of training dynamics include the information bottleneck, double descent, neural collapse, and grokking \citep{shwartz2017opening,nakkiran2021deep,papyan2020prevalence,power2022grokking}. Related approaches to characterizing representations include intrinsic dimension \citep{ansuini2019intrinsic}, representation topology divergence \citep{pmlr-v162-barannikov22a}, and CKA \citep{kornblith2019similarity}, alongside further topological analysis \citep{ballester2026topological} and ricci curvature analysis \citep{sun2025ricciflowrec}. Our experiments examine persistence alongside activation scale and local geometry to understand which aspects of the response to noisy supervision each measure captures.

\section{Discussion and Conclusion}
\label{sec:conclusion}

Persistent homology reveals representation differences under noisy supervision that survive uniform-scale controls. On the fixed training probe, corrupted subsets develop higher persistence than clean subsets at a checkpoint where predictions for most corrupted examples still match their original labels. Across pipelines, crossover depends on the distance construction, while activation norm provides a stronger average in-sample fit than raw persistence for noise-rate calibration. These results
distinguish estimating noise amount from understanding how it affects learned representations. Within-model comparisons, scale controls, and label-fitting measurements provide a practical approach to studying this distinction: they reveal where geometric differences arise, which survive uniform rescaling, and how they relate to the predictions made during training.

\section*{Acknowledgements and Disclosures}
This research received no specific funding. The authors declare no competing interests.
\clearpage
\bibliographystyle{plainnat}
\bibliography{ref}
\newpage
\appendix

\section{Impact Statement}
\label{app:impact}
Understanding how noisy supervision changes representations may help researchers audit datasets and compare training procedures. Persistence describes groups of examples, so interpreting it as evidence of individual annotation errors risks confusing effects of the training pipeline with label quality. Using it alone to filter data could remove valid examples, including those from underrepresented groups. Data selection should therefore combine the diagnostic with annotation review and an assessment of its effects across groups.

\section{Training Hyperparameters}
\label{app:hyperparams}

\paragraph{Symmetric uniform relabelling.}
A fraction $\mathrm{sf}$ of training examples is selected for uniform relabelling over all classes, including the
original class. Thus $\mathrm{sf}$ denotes the selection fraction rather than the realised fraction of changed labels.
In the fixed training probe, corruption membership is determined by whether the assigned label differs from the original label.
 
\begin{table}[h]
\centering
\caption{Training hyperparameters for the architectures used in the main experiments.}
\small
\begin{tabular}{@{}lcccccc@{}}
\toprule
Architecture & Optimizer & LR & Schedule & Weight Decay & Batch & Epochs \\
\midrule
MLP & Adam & 0.001 & Cosine & 0 & 128 & 50 \\
ResNet-18/50/101 & SGD+mom. & 0.1 & Cosine & $5 \times 10^{-4}$ & 128 & 50/100 \\
ViT-Tiny/Large & AdamW & 0.001 & Cosine & 0.01 & 128 & 50/100 \\
DistilBERT & AdamW & $2 \times 10^{-5}$ & Lin.\ warmup & 0.01 & 32 & 50 \\
\bottomrule
\end{tabular}
\end{table}

Synthetic 2D setup. We train a 2-hidden-layer MLP with 64 ReLU units per layer (input dimension 2, output dimension 2) for 100 epochs using Adam with learning rate 0.01. Boundary points are defined as points satisfying $|x_0| < 1.0$, that is, within distance 1.0 of the optimal vertical separator between the two clusters. The two 2D Gaussian clusters are centred at $(-2,0)$ and $(+2,0)$ with unit variance, and we use 20 seeds per flip rate. The flip fraction is defined over the near-boundary points, not over the entire training set.

\paragraph{Persistence conventions.}
The relative lifetime filter in Eq.~\ref{eq:phi} is used for the normalisation, slope, crossover, training-probe,
NEAR/FAR, revised classwise, pair-distance, synthetic, ViT noise-type, exploratory class-pair, ImageNette,
Tiny-ImageNet, and CIFAR-10N analyses. Grokking, TGCS, and the original large-scale corruption and pretrained analyses sum all finite lifetimes. The $H_0$ summary excludes the infinite bar. Ripser uses coefficients in $\mathbb{Z}/2\mathbb{Z}$. Ratios are set to zero when their denominators are zero.


\paragraph{Normalisation.}
In the normalisation grid, each transformation is applied to the selected 300-point subsample. Unit normalisation divides each activation vector by its Euclidean norm. Z-score standardisation subtracts each feature's mean and divides by its standard deviation, computed over the subsample with divisor $n$. RMS rescaling centres the cloud and divides by the root mean squared norm of the centred vectors. All three denominators are bounded below by $10^{-12}$. Cosine dissimilarities are computed directly from the selected activations.

\paragraph{Classwise protocols.}
The revised protocol groups saved test activations by true labels and sums persistence across eligible classes.
It uses a budget of 50 points per class, includes classes with at least 10 examples, and skips classwise analysis
for datasets with more than 20 classes. This protocol was introduced after the original results were available.

The original ViT protocol groups activations by noisy labels from an augmented training loader. It includes classes with at least 30 examples and samples up to 300 points per class without a fixed subsampling seed. The two protocols therefore differ in data source, class labels, eligibility, and sampling budget.

\paragraph{Statistical summaries.}
Table~\ref{tab:training_probe} reports ratios of seed medians, with tests applied to paired values within each run. For Table~\ref{tab:noise_seed_controls}, the three per-setting ratios are reduced to a median within each seed. These seed-level medians are tested against one using an exact two-sided Wilcoxon signed-rank test. Pooled analyses across seeds and noise settings are descriptive, as observations sharing a seed are dependent. Slope intervals use 2,000 bootstrap resamples of runs without clustering by seed. Benjamini--Hochberg correction is applied across the cells within each reported slope analysis.

\paragraph{Gaussian controls.}
Gaussian controls are fitted using the class-conditional means and covariances of the observed subset.
Grid controls use the 300-point subsample and five Gaussian draws. Training-probe controls use the matched
subset grouped by true labels and ten draws, with persistence evaluated on 300 points. The comparison between subsets tests the difference in excess persistence relative to their fitted controls.

 
\section{Subsampling Sensitivity}
\label{app:subsampling}

Table~\ref{tab:subsampling} compares persistence rankings from subsamples with those from a 2,000-point reference across 433 configurations. At $n=300$, the Spearman correlation is $0.973$, with a runtime of $0.058$\,s per call compared with $2.944$\,s for the reference. The mean subsampling coefficient of variation is $0.022$, with a maximum of $0.086$.

\begin{table}[h]
\centering
\caption{Subsampling sensitivity across 433 configurations. $\rho$ measures rank agreement with the 2,000-point reference. Dashes indicate unreported runtimes.}
\label{tab:subsampling}
\small
\begin{tabular}{@{}lcccccccc@{}}
\toprule
$n$ & 50 & 100 & 200 & 300 & 500 & 750 & 1000 & 1500 \\
\midrule
$\rho$ & 0.908 & 0.922 & 0.960 & 0.973 & 0.979 & 0.982 & 0.986 & 0.998 \\
Time (s) & --- & --- & --- & 0.058 & --- & --- & 0.612 & --- \\
\bottomrule
\end{tabular}
\end{table}

A separate CIFAR-10/ResNet-18 experiment at epoch~10 varies the global point count over $n\in\{100,200,300,500,1000,2000\}$. Across three noise levels and five seeds, the Spearman correlation with noise level is $-0.850$ at $n=100$ and $-0.945$ at every tested size from $n=200$ onward. The repeated correlation describes the strength of the rank association, while persistence magnitudes and sampling variability may still differ. This experiment measures association with noise level, whereas Table~\ref{tab:subsampling} measures agreement with a larger reference cloud. Both analyses vary point count while leaving sensitivity to the lifetime cutoff for separate evaluation.

\section{Raw Persistence Associations over Training}
\label{app:fading}

Table~\ref{tab:noise_associations} reports Spearman correlations between raw persistence and noise level at epochs~10 and~50. These correlations describe how noise conditions are ordered at each checkpoint. Section~\ref{sec:crossover} examines when noisy persistence overtakes its clean reference under different distance constructions. The original CIFAR-100/ResNet-18 row is omitted following recomputation, with the updated crossover analysis reported in that section.

\begin{table}[h]
\centering\small
\caption{Spearman correlations between noise level and raw persistence at epochs~10 and~50. Changes in correlation describe changes in noise ordering, whereas crossover concerns noisy persistence relative to a clean reference. Values use the raw, 300-point reanalysis consistently across pipelines.}
\label{tab:noise_associations}
\begin{tabular}{@{}llcc@{}}
\toprule
Dataset & Architecture & $\rho_{10}$ & $\rho_{50}$ \\
\midrule
CIFAR-10 & ResNet-18 & $-0.969$ & $-0.225$ \\
MNIST & MLP & $-0.931$ & $-0.318$ \\
CIFAR-10 & ViT-Tiny & $-0.962$ & $-0.962$ \\
FashionMNIST & MLP & $-0.954$ & $-0.876$ \\
AG News & DistilBERT & $+0.870$ & $+0.870$ \\
\bottomrule
\end{tabular}
\end{table}

\section{Large-Scale Results}
\label{app:largescale}
The original extended CIFAR-100/ResNet-50 experiment reports a crossover at epochs 60--70. The corresponding original ResNet-18 crossover has been superseded by the no-crossover recomputation, so these observations do not establish a monotone delay of crossover with depth. A matched analysis using the same extraction, normalisation, and checkpoint protocol is needed for that comparison.

\section{Negative Results}
\label{app:negative}

\paragraph{ACRP as a PH proxy.}
Activation Covariance Rank Proxy (ACRP) is the effective rank of the activation sample covariance matrix, defined as $(\mathrm{tr}(\Sigma))^2 / \|\Sigma\|_F^2$, where $\Sigma$ is the sample covariance of the penultimate-layer activations. This quantity, also known as the participation ratio, captures how activation variance is distributed across feature dimensions, with higher values indicating more uniform spread. We tested ACRP as a Ripser-free proxy for monitoring topological complexity in real time. It correlates with $\Phi$ on some datasets, including CIFAR-10/ResNet-18 ($\rho = 0.831$) and CIFAR-100 runs ($\rho=0.504$, $n=168$), but not on others, including FashionMNIST ($\rho = 0.002$) and MNIST ($\rho = -0.321$), so it is not a universal proxy for persistent homology.

\paragraph{$\Phi$ peak as an accuracy predictor.}
Peak $\Phi$ is not a reliable predictor of final accuracy. The overall correlation between peak height and final accuracy is only $\rho = 0.27$, although it rises to $\rho = 0.56$ at sf$=0.5$. Stopping at the $\Phi$ peak improves performance by $4.2\%$ at sf$=0.5$ but reduces it by $4.1\%$ at sf$=0.0$. 

\section{TGCS Algorithm}
\label{app:tgcs_algo}

Pseudocode for the topology-guided checkpoint selection (TGCS) procedure used in the appendix analysis.

\begin{quote}
\textbf{Topology-Guided Checkpoint Selection (TGCS)}\label{alg:tgcs}

\textit{Inputs:} training procedure, validation set $\mathcal V$, subsample size $n=300$, and final epoch $T$.
\begin{enumerate}[nosep,leftmargin=1.5em]
\item Train for $T$ epochs, computing $\Phi(t)$ at each checkpoint.
\item Let $t_{\mathrm{peak}}=\operatorname*{arg\,max}_t\Phi(t)$.
\item Retain the checkpoints at $t_{\mathrm{peak}}$ and $T$.
\item Evaluate both on $\mathcal V$ and return the one with higher validation accuracy.
\end{enumerate}
\end{quote}
 
\section{Compute Budget}
\label{app:compute}
 

\begin{table}[h]
\centering
\caption{Training used the GPU partition, including A100 and older GPU models; persistent-homology computations and response-period analyses used CPUs. GPU-hours and CPU-hours are reported separately.}
\small
\begin{tabular}{@{}lrr@{}}
\toprule
Component & GPU-hours & CPU-hours \\
\midrule
Original training (900+ runs)  & $\sim$2{,}400 & --- \\
Grokking experiments (12 jobs) & $\sim$48      & --- \\
Realistic noise (39 jobs)      & $\sim$195     & --- \\
TGCS evaluation (40 jobs)      & $\sim$120     & --- \\
PH computation                 & ---          & $\sim$50 \\
\bottomrule
\end{tabular}
\end{table}

\section{Regularisation Methods}
\label{app:regulariser}


We compare six training methods on CIFAR-10/ResNet-18: the baseline, dropout, label smoothing, high weight decay, topology-guided training (TGT), and validation-based early stopping. TGT uses $\Phi$ to guide learning-rate and weight-decay adjustments. Results are averaged over noise levels. At epoch~50, label smoothing and high weight decay reduce $\Phi$ by approximately 50\% relative to the baseline, whereas TGT increases it. Test accuracy does not follow a monotone ordering with persistence across these methods (Table \ref{tab: regularisation}).
 
\begin{table}[h]
\centering
\caption{Regularisation methods on CIFAR-10/ResNet-18 (averaged over noise levels).}
\label{tab: regularisation}
\small
\begin{tabular}{@{}lcc@{}}
\toprule
Method & $\Phi$ (ep50) & Test accuracy \\
\midrule
Baseline           & 357.0 & 52.5\% \\
Dropout            & 351.9 & 52.3\% \\
Label smoothing    & 174.0 & 52.8\% \\
Weight decay (high) & 163.6 & 46.8\% \\
TGT                & 618.6 & 50.1\% \\
Val.\ early stop   & ---   & 51.4\% \\
\bottomrule
\end{tabular}
\end{table}

\section{Grokking Results}
\label{app:grokking}


The experiment evaluates two-layer Transformers on modular addition and multiplication with primes 97 and 113, using three seeds per task and prime. For each run, $\Delta\Phi_{\mathrm{grok}}$ is the mean test-set persistence over checkpoints with training accuracy $\geq0.99$ and test accuracy $\geq0.90$, minus the mean over checkpoints with training accuracy $\geq0.99$ and test accuracy $<0.50$ (Table~\ref{tab:grokking_results}). Persistence is computed on 300 test points by summing all finite $H_1$ lifetimes without the relative filter. Ten runs show a decrease, and two show increases of $3.7$ and $4.5$.

\begin{table}[h]
\centering
\small
\caption{Changes in raw persistence across twelve grokking runs. The mean change is $-35.2$ across all runs and $-43.1$ among the ten decreasing runs.}
\label{tab:grokking_results}
\begin{tabular}{@{}llrrc@{}}
\toprule
Task & Prime & Seed & $\Delta\Phi_{\mathrm{grok}}$ & Decreases? \\
\midrule
Addition & 97  & 42  & $-31.2$ & Yes \\
Addition & 97  & 123 & $-46.9$ & Yes \\
Addition & 97  & 456 & $-25.7$ & Yes \\
Addition & 113 & 42  & $+3.7$  & No \\
Addition & 113 & 123 & $-29.4$ & Yes \\
Addition & 113 & 456 & $-64.3$ & Yes \\
Multiplication & 97  & 42  & $-7.4$  & Yes \\
Multiplication & 97  & 123 & $-90.1$ & Yes \\
Multiplication & 97  & 456 & $-17.3$ & Yes \\
Multiplication & 113 & 42  & $-82.1$ & Yes \\
Multiplication & 113 & 123 & $-36.5$ & Yes \\
Multiplication & 113 & 456 & $+4.5$  & No \\
\bottomrule
\end{tabular}
\end{table}

\begin{figure}[tbp]
\centering
\includegraphics[width=0.96\textwidth]{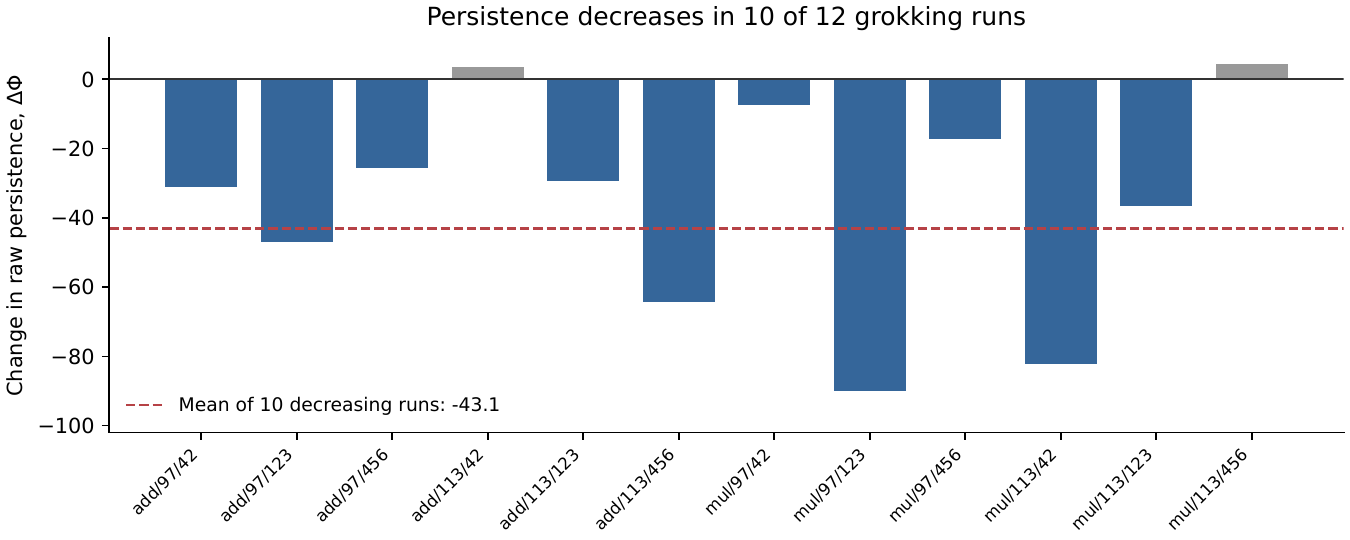}
\caption{Changes in raw persistence across the grokking transition for all twelve runs. Persistence decreases in ten runs and increases by $3.7$ and $4.5$ in the remaining two. The mean change is $-35.2$ across all runs and $-43.1$ among the ten decreasing runs. The dashed red line marks the latter mean.}
\label{fig:grokking}
\end{figure}

\section{Calibration Targets}
\label{app:noise_est}



Table~\ref{tab:noise_compare} uses five pipelines: CIFAR-10/ResNet-18, CIFAR-10/ViT-Tiny, MNIST/MLP, FashionMNIST/MLP, and CIFAR-100/ResNet-18. Within each pipeline, ordinary least squares with an intercept predicts the nominal noise setting from one epoch-10 representation statistic, using individual runs as observations. Fitting and evaluation use the same observations; the reported $R^2$ values are in-sample. The statistics are mean activation norm, raw persistence, point-cloud diameter, unit-normalised persistence, and $\Phi_{H_1}/\Phi_{H_0}$. Ranges and means in Table~\ref{tab:noise_compare} summarize the five pipeline-specific $R^2$ values. Mean activation norm and raw persistence achieve mean $R^2$ values of $0.936$ and $0.908$. After excluding the fully randomised condition, the corresponding means are $0.892$ and $0.729$. These fits require runs at known noise settings and do not establish held-out calibration accuracy.

\section{\texorpdfstring{$\Phi$}{Phi}-Guided Adaptive Label Smoothing}
\label{app:adaptive}

We test whether the early noise estimate can be used to drive adaptive regularisation. At epoch~10, we compute $\hat{s}$ from $\Phi(10)$ and then set the label-smoothing strength to $\hat{s} \times 0.8$ for the remaining epochs. Results on CIFAR-10/ResNet-18 (4 noise levels, 3 seeds each) are shown below.
 
\begin{table}[h]
\centering
\caption{$\Phi$-guided adaptive vs.\ fixed label smoothing (0.1). Bold = best per row.}
\small
\begin{tabular}{@{}ccccc@{}}
\toprule
sf & Baseline & Fixed (0.1) & Adaptive & Winner \\
\midrule
0.00 & \textbf{86.92} & 86.74 & 86.77 & Baseline \\
0.10 & 84.96 & \textbf{85.62} & 85.25 & Fixed \\
0.25 & 82.64 & 83.16 & \textbf{83.47} & Adaptive \\
0.50 & 77.00 & \textbf{77.50} & 77.28 & Fixed \\
\midrule
Mean & 82.88 & \textbf{83.26} & 83.19 & Fixed \\
\bottomrule
\end{tabular}
\end{table}
 
Adaptive smoothing wins at sf$=$0.25 ($+0.31$ percentage points over fixed).
However, it does not outperform fixed smoothing on average ($-0.07$ percentage points using the displayed means).
The reported accuracy differences do not isolate whether calibration error or the smoothing rule limits performance.
We include this as a proof-of-concept: the topological signal can be used in a regularisation rule, but this rule does not beat the fixed baseline on average.
 
\section{TGCS Results}
\label{app:tgcs_results}

Table \ref{tab:tgcs result} presents an illustrative subset of the TGCS results. Here $\delta$ is the difference in test accuracy between the persistence-peak and final checkpoints, clipped at zero. It measures the positive gain of the peak checkpoint over the final checkpoint. Evaluating the benefit of TGCS as a selection procedure requires comparison with validation-based early stopping using the same checkpoint budget.


\begin{table}[h]
\centering
\caption{TGCS results (representative subset of 40 runs).
$\delta=100\max(0,a_{\mathrm{peak}}-a_{\mathrm{final}})$,
where $a_{\mathrm{peak}}$ and $a_{\mathrm{final}}$ are test
accuracies on a 0--1 scale at the $\Phi$-peak and final
checkpoints, respectively. Reported clean-data and
ViT-collapsed entries have $\delta=0$.}
\label{tab:tgcs result}
\small
\begin{tabular}{@{}llcccc@{}}
\toprule
Dataset & Arch & sf & Seed & Final acc
& $\delta$ (percentage points) \\
\midrule
CIFAR-100    & ResNet & 0.5 & 42  & 0.3842 & $+10.0$ \\
CIFAR-100    & ResNet & 0.5 & 123 & 0.3904 & $+11.0$ \\
FashionMNIST & MLP    & 0.5 & 42  & 0.7265 & $+5.0$  \\
FashionMNIST & MLP    & 0.5 & 123 & 0.7263 & $+3.6$  \\
FashionMNIST & MLP    & 0.5 & 456 & 0.7285 & $+2.5$  \\
MNIST        & MLP    & 0.5 & 42  & 0.6804 & $+3.3$  \\
CIFAR-10     & ResNet & 0.5 & 42  & 0.7407 & $+1.3$  \\
\midrule
\multicolumn{4}{@{}l}{All clean (sf$=$0.0)}
& --- & $0.0$ \\
\multicolumn{4}{@{}l}{ViT-Tiny (collapsed)}
& --- & $0.0$ \\
\bottomrule
\end{tabular}
\end{table}
 
\section{Depth Ablation}
\label{app:depth}
The original depth ablation used activations extracted during training, whereas the canonical tables used saved activations and a different computation pipeline. Its absolute values cannot be interchanged with the canonical reanalysis. Because the CIFAR-100/ResNet-18 baseline was subsequently recomputed and no longer crosses, the original depth figure and the $1.5\times$--$6.4\times$ ratio comparison are withheld pending reconciliation; they do not currently support a general claim that depth delays folding.

\section{ViT Per-Class PH Investigation}
\label{app:vit}
 
We reanalyse all 280 saved ViT-Tiny activation files using per-class persistent homology. This is a secondary classwise analysis; its aggregate magnitude differs from a single global computation because its grouping and point budget differ. 
\begin{table}[h]
\centering
\caption{ViT-Tiny/CIFAR-10: per-class $\Phi$ by noise level (mean over 5 seeds). The classwise sum is larger than the global sum and has noise correlation $\rho \leq -0.93$; unequal point budgets preclude interpreting the magnitude ratio as diagnostic improvement.}
\small
\begin{tabular}{@{}ccccc@{}}
\toprule
sf & Per-class $\Phi$ & Global ($n{=}300$) & Global ($n{=}1000$) & Epoch \\
\midrule
0.00 & 260.3 & 32.1 & 118.8 & 10 \\
0.25 & 143.0 & 18.5 & 60.3 & 10 \\
0.50 & 118.5 & 13.5 & 48.9 & 10 \\
1.00 & 0.0 & 0.0 & 0.0 & 10 \\
\midrule
0.00 & 403.4 & 66.1 & 229.0 & 50 \\
0.25 & 160.5 & 25.8 & 82.3 & 50 \\
0.50 & 168.6 & 22.2 & 74.5 & 50 \\
1.00 & 0.0 & 0.0 & 0.0 & 50 \\
\bottomrule
\end{tabular}
\end{table}

Clean classwise persistence rises from about 260 to 403. Both global and classwise values retain a noise association, and the original classwise analysis reports $\rho\leq-0.93$, $p<10^{-9}$. A larger aggregate total cannot on its own distinguish inter-class interference from the effect of computing and summing PH over more points. These results therefore establish sensitivity to grouping and sample budget rather than a unique explanation of the original weak signal.

\section{Tiny-ImageNet Scaling}
\label{app:tinyimagenet}

We train ResNet-18 on Tiny-ImageNet (200 classes, 64$\times$64, 100K train images) with 4 noise levels and 5 seeds each (20 runs total, 50 epochs). For this experiment, we save a larger activation set per checkpoint and use 50 classes $\times$ 300 points for per-class PH analysis.
 
\begin{table}[h]
\centering
\caption{Tiny-ImageNet/ResNet-18: global vs.\ per-class $\Phi$ by noise level (mean over 5 seeds). The larger classwise aggregate discriminates noise with $\rho = -0.969$ ($p < 10^{-12}$) at epochs 10 and 25.}
\small
\begin{tabular}{@{}ccccc@{}}
\toprule
sf & Global ($n{=}300$) & Per-class & Val Acc & Epoch \\
\midrule
0.00 & 84.4 & 4{,}035.2 & 31.0\% & 10 \\
0.25 & 82.3 & 3{,}311.4 & 25.4\% & 10 \\
0.50 & 54.3 & 2{,}554.4 & 19.2\% & 10 \\
1.00 & 0.6 & 26.5 & 0.5\% & 10 \\
\midrule
0.00 & 165.6 & 5{,}332.6 & 50.3\% & 50 \\
0.25 & 94.5 & 4{,}638.8 & 35.8\% & 50 \\
0.50 & 75.7 & 4{,}934.8 & 21.3\% & 50 \\
1.00 & 4.1 & 222.7 & 0.5\% & 50 \\
\bottomrule
\end{tabular}
\end{table}
 
\textbf{Interpretation.} Global PH at $n=300$ separates the reported clean and noisy conditions. Classwise PH uses 50 classes with 300 points each (15,000 points in total), so its larger aggregate is not a matched-budget comparison. Its noise correlation is $\rho=-0.969$ ($p<10^{-12}$, $n=20$ runs) at epochs 10 and 25 and $-0.752$ at epoch 50. Global clean persistence rises from 84.4 to 165.6. The table supports a persistent clean trajectory, without establishing a universal rule that increasing clean persistence precludes crossover.

\section{Prospective Crossover Test on ResNet-34}
\label{app:r34}

CIFAR-10/ResNet-34 provides a prospective test of a crossover prediction based on clean training.
This configuration was not used to derive C1/C2 and follows the same training recipe as ResNet-18.

Across three initial clean runs, mean raw persistence rose from $7.4$ at epoch~1 to $28.2$ at epoch~5, peaked at $32.9$ at epoch~35, and ended at $31.7$ at epoch~50 (Table~\ref{tab:r34_trajectory}). Clean persistence thus remained elevated throughout the later checkpoints. This trajectory motivated a no-crossover prediction, recorded before noisy training began.

Subsequent training used three seeds at each of $\mathrm{sf}=0.25$ and $\mathrm{sf}=0.5$. All six noisy runs remained below the clean reference at every sampled checkpoint, consistent with the prediction. The contrast with ResNet-18 on CIFAR-10 illustrates variation in crossover behavior across architectures under the reported training recipe.

\begin{table}[h]
\centering
\caption{Raw persistence on CIFAR-10/ResNet-34, averaged over three seeds per noise setting. The clean trajectory motivated a no-crossover prediction before noisy training began. All six noisy runs remained below the clean reference at every sampled checkpoint.}
\label{tab:r34_trajectory}
\small
\begin{tabular}{@{}cccc@{}}
\toprule
Epoch & $\mathrm{sf}=0$ (clean) & $\mathrm{sf}=0.25$ & $\mathrm{sf}=0.5$ \\
\midrule
1  & 7.4  & 2.6  & 0.3  \\
5  & 28.2 & 16.7 & 6.0  \\
10 & 28.2 & 16.8 & 10.7 \\
25 & 29.5 & 16.5 & 12.8 \\
35 & 32.9 & 17.1 & 12.5 \\
50 & 31.7 & 18.6 & 16.4 \\
\bottomrule
\end{tabular}
\end{table}

\section{ViT Noise-Type Test (Asymmetric vs Instance-Dependent)}
\label{app:vit_noisetype}
 

ViT-Tiny is trained on CIFAR-10 with cyclic asymmetric noise and pixel-distance-based instance-dependent noise at nominal rates of $0.2$ and $0.4$, using five seeds per condition. Cyclic asymmetric noise maps class $c$ to $(c+1)\bmod 10$, a different mapping from the Patrini construction. Instance-dependent noise uses pixel-space distances from
the true-class centroid and assigns labels uniformly, allowing the original class to be redrawn. The resulting fractions of changed labels are approximately $0.134$ and $0.267$, respectively. Persistence is computed within each class using a budget of 300 points per class. These results come from a separate protocol from the ResNet-18 relabelling experiment.
 
\begin{table}[h]
\centering
\caption{Per-class $\Phi$ on ViT-Tiny under asymmetric vs.\ instance-dependent noise (mean $\pm$ std over 5 seeds). The two noise types produce no statistically detectable difference in raw persistence at epoch 10 ($p > 0.3$) but different raw persistence at epoch 50 ($p \leq 0.002$).}
\small
\begin{tabular}{@{}lccccc@{}}
\toprule
Epoch & Rate & Asymmetric $\Phi$ & Instance $\Phi$ & Ratio & $t$-test $p$ \\
\midrule
10 & 0.2 & $821.1 \pm 42.3$ & $796.7 \pm 43.7$ & 0.97 & 0.395 \\
10 & 0.4 & $777.9 \pm 32.9$ & $774.8 \pm 31.1$ & 1.00 & 0.885 \\
25 & 0.2 & $660.5 \pm 58.3$ & $610.9 \pm 24.5$ & 0.92 & 0.117 \\
25 & 0.4 & $662.0 \pm 28.9$ & $711.1 \pm 37.1$ & 1.07 & 0.048 \\
50 & 0.2 & $426.6 \pm 29.5$ & $505.4 \pm 27.1$ & \textbf{1.18} & \textbf{0.002} \\
50 & 0.4 & $485.8 \pm 29.5$ & $717.3 \pm 49.0$ & \textbf{1.48} & $\mathbf{< 0.001}$ \\
\bottomrule
\end{tabular}
\end{table}

\textbf{Interpretation.} For ViT-Tiny, no difference in raw persistence between the two noise types is detected at epoch~10. At epoch~50, instance-dependent noise yields higher raw persistence at both rates ($p=0.002$ and $p<0.001$,
respectively, from independent two-sample $t$-tests with five seeds per arm). The ResNet-18 comparison in Section 5 uses different corruption generators and seed-level paired tests. Its random-relabeling and Patrini arms differ in realised corruption rate.


\section{Global and Classwise Persistence on AG News}
\label{app:distilbert}

Saved DistilBERT/AG News activations are analysed using global PH with 300 and 1,000 points and classwise PH
with 300 points from each of four classes. Table~\ref{tab:distilbert_ph} reports raw persistence
under clean training and $\mathrm{sf}=0.5$, averaged over five seeds.

All three analyses show a positive Spearman correlation with noise level ($\rho=0.870$, $p<0.002$). At epoch~10, the classwise sum increases from $103.7$ under clean training to $371.4$ at $\mathrm{sf}=0.5$,
a noisy/clean ratio of approximately $3.6$. This sum combines four separate 300-point computations,
with a different point budget and aggregation from global PH.

\begin{table}[h]
\centering
\caption{DistilBERT/AG News: $\Phi$ at clean vs.\ sf$=$0.5 (mean over 5 seeds). All three $\Phi$ variants show $\rho = +0.870$ ($p < 0.002$) with a consistent \emph{sign reversal} compared to vision: noise raises raw total persistence; scale and loop count are not separated by this table.}
\label{tab:distilbert_ph}

\small
\begin{tabular}{@{}cccccc@{}}
\toprule
Epoch & sf & Global ($n{=}300$) & Per-class & Global ($n{=}1000$) & Noisy/Clean Ratio \\
\midrule
10 & 0.00 & 18.1 & 103.7 & 56.1 & --- \\
10 & 0.50 & 53.1 & 371.4 & 224.5 & $2.9\times$ / $3.6\times$ / $4.0\times$ \\
\midrule
25 & 0.00 & 15.7 & 83.0 & 43.9 & --- \\
25 & 0.50 & 56.7 & 354.2 & 178.3 & $3.6\times$ / $4.3\times$ / $4.1\times$ \\
\midrule
50 & 0.00 & 12.2 & 77.6 & 40.2 & --- \\
50 & 0.50 & 45.3 & 285.4 & 143.8 & $3.7\times$ / $3.7\times$ / $3.6\times$ \\
\bottomrule
\end{tabular}
\end{table}

Clean mean persistence declines across the reported checkpoints, while noisy means remain above their
clean references in all three analyses. Noisy persistence also declines in the classwise and global $n=1000$ analyses. For global $n=300$, it rises slightly at epoch~25 before declining at epoch~50. The positive noise association contrasts with the early negative associations in the evaluated vision settings, motivating calibration within each pipeline.

\section{Sign-Reversal Diagnosis: Transformer on Vision}
\label{app:transformer_vision}
 
To help distinguish modality from architecture as a possible driver of the DistilBERT sign reversal, we train a small Transformer (4 layers, embed dim 128, 4 heads, $8{\times}8$ patches) on CIFAR-10 with 3 noise levels ($\text{sf} \in \{0.0, 0.25, 0.5\}$) and 5 seeds each (15 runs total).

\begin{table}[h]
\centering
\caption{Small Transformer on CIFAR-10: $\Phi$ by noise level and epoch (mean over 5 seeds, $n=15$ total). Noise \emph{decreases} $\Phi$ at all epochs tested, matching the vision pattern on ResNet/ViT, opposite to DistilBERT/text.}
\small
\begin{tabular}{@{}cccccc@{}}
\toprule
Epoch & sf & Global $\Phi$ & Per-class $\Phi$ & Val Acc & $n$ \\
\midrule
10 & 0.00 & 130.5 & 982.3 & 59.4\% & 5 \\
10 & 0.25 & 79.6 & 731.2 & 52.6\% & 5 \\
10 & 0.50 & 62.5 & 606.7 & 44.8\% & 5 \\
\midrule
25 & 0.00 & 129.4 & 911.2 & 68.3\% & 5 \\
25 & 0.25 & 73.6 & 599.9 & 63.0\% & 5 \\
25 & 0.50 & 71.8 & 663.4 & 55.5\% & 5 \\
\midrule
50 & 0.00 & 122.9 & 815.2 & 72.7\% & 5 \\
50 & 0.25 & 61.8 & 524.6 & 67.7\% & 5 \\
50 & 0.50 & 68.1 & 629.4 & 60.4\% & 5 \\
\bottomrule
\end{tabular}
\end{table}

\textbf{Result.} The Spearman correlation between $\Phi$ and noise level is negative throughout training. At epoch~10, global and per-class PH both give $\rho = -0.945$ ($p < 0.001$). At epoch~25, the correlations are $\rho = -0.775$ ($p = 0.001$) for global PH and $\rho = -0.510$ ($p = 0.05$) for per-class PH. At epoch~50, they are $\rho = -0.548$ ($p = 0.034$) and $\rho = -0.491$ ($p = 0.06$), respectively.

\textbf{Interpretation.} The sign is clearly negative for a vision Transformer at the diagnostic epoch ($p < 0.001$, $n=15$), in contrast to the positive sign observed on DistilBERT/AG News. The association weakens over training but remains negative, with global correlations significant at all three reported checkpoints. This comparison shows that a Transformer on vision need not share the positive association observed for DistilBERT on text. It does not isolate modality from the many differences between the two training pipelines.




\section{Additional normalisation and Pair Controls}
\label{app:pair_controls}

\paragraph{Original exploratory class-pair experiment.}

The exploratory CIFAR-10/ResNet-18 experiment applies a nominal pairwise flip parameter of 0.5, with ten seeds per condition. Pairs are processed sequentially on already modified labels and reuse classes. The realised changed-label fraction is approximately 42.5\%, with unequal corruption across classes. The first set, selected for a high noisy/clean persistence ratio, contains \{horse$\leftrightarrow$truck, automobile$\leftrightarrow$truck, automobile$\leftrightarrow$dog, frog$\leftrightarrow$ship, cat$\leftrightarrow$dog\}. The comparison set contains \{airplane$\leftrightarrow$automobile, airplane$\leftrightarrow$ship, bird$\leftrightarrow$ship, deer$\leftrightarrow$frog, bird$\leftrightarrow$truck\}.

\begin{figure}[t]
\centering
\includegraphics[width=0.80\textwidth]{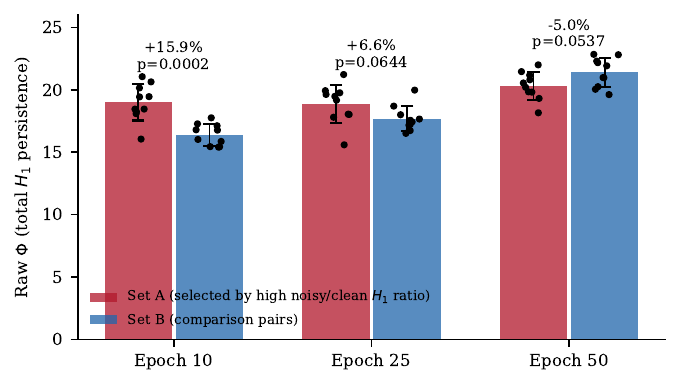} 
\caption{Exploratory class-pair comparison on CIFAR-10/ResNet-18. Set A was selected using a high noisy/clean persistence ratio; Set B contains comparison pairs. Bars show means, error bars show sample standard deviations, and dots show individual runs across ten seeds per condition. Reported $p$-values use Welch's two-sample $t$-test. The nominal pairwise flip parameter is 0.5; sequential flips over overlapping pairs change approximately 42.5\% of labels,
unevenly across classes. These conditions are distinct from the disjoint, centroid-selected NEAR and FAR comparison.}
\label{fig:classpair}
\end{figure}

At epoch~10, raw persistence is $19.0\pm1.5$ in the first condition and $16.4\pm0.9$ in the comparison condition, a relative difference of $15.9\%$ ($t=4.87$, $p=0.0002$, Welch’s two-sample $t$-test). The reported dispersions are sample standard deviations across seeds. The difference decreases to $6.6\%$ at epoch~25 ($p=0.0644$) and $-5.0\%$ at epoch~50 ($p=0.054$), as shown in Figure~\ref{fig:classpair}. Because persistence informed selection of the first set, this experiment is exploratory rather than an independent test of the selection criterion. Unlike the centroid-based NEAR and FAR conditions, these sets contain overlapping class pairs.

\paragraph{Pair-distance analysis.}

The pair-distance analysis uses $sf=0.5$. For each class pair, clean centroid distance and the noisy/clean
persistence ratio are first averaged across seeds; Spearman correlation is then computed across pairs. Both CIFAR-10 analyses use all 45 unordered pairs and five seeds. The CIFAR-100 analysis uses 60 pairs selected with RandomState(0).

\begin{table}[h]
\centering\small
\caption{Pair-distance correlations reported in the response-period analysis. Both CIFAR-10 configurations have $p<0.0001$ at epochs~25 and~50. Correlations are computed across pairs after seed averaging at $sf=0.5$.}
\label{tab:pair_controls}
\begin{tabular}{@{}lrrrr@{}}
\toprule
Pipeline & Pairs & Epoch 10 & Epoch 25 & Epoch 50 \\
\midrule
CIFAR-10 / ResNet-18 & 45 & $-0.471$ & 0.823 & 0.867 \\
CIFAR-10 / ViT-Tiny & 45 & 0.104 & 0.664 & 0.614 \\
CIFAR-100 / ResNet-18 & 60 & $-0.172$ & $-0.119$ & 0.413 \\
\bottomrule
\end{tabular}
\end{table}

\paragraph{Alternative summaries.}
Scale-invariant bar count correlates with noise at $\rho=-0.799$ on CIFAR-10/ResNet-18 and $-0.830$ on CIFAR-10/ViT-Tiny (both $p<0.001$). The corresponding values on CIFAR-100, FashionMNIST, and MNIST are $-0.411$, $0.000$, and $0.361$, respectively (none significant). Thus loop count carries some of the signal, but does not replace lifetime-based summaries across these settings. The relative threshold is fixed throughout. A full threshold sensitivity experiment is not reported.

\paragraph{Classwise normalised analysis.}
For CIFAR-10/ResNet-18, the response-period classwise scale-invariant ratio has noise correlation $\rho=-0.737$ at epoch~10 ($p=0.002$) and $0.794$ at epoch~50 ($p<0.001$). This documents a change in noise ordering. A correlation sign change alone does not give the first clean--noisy crossing time; that requires the trajectories themselves.

\paragraph{Seed-level relabelling comparisons.}
Each seed-level ratio is the median of the three per-setting ratios. At epochs~10, 25, and~50, respectively, the random-relabeling arm has higher $\Phi_{H_1}/\Phi_{H_0}$ than the Patrini arm in 7/8, 6/8, and 8/8 seeds, with reported paired Wilcoxon $p=0.0156,0.0391,0.0078$. Mean norm is lower in 8/8 seeds at each checkpoint (median ratios 0.608, 0.580, and 0.702; each $p=0.0078$). Diameter is likewise lower in 8/8 seeds, with ratios 0.517, 0.448, and 0.790. These results and the cosine analysis show that activation scale and normalised persistence rank the two conditions in opposite orders. The comparison involves differences in both realised corruption rate and relabelling structure.


\clearpage
\end{document}